\pdfoutput=1
\documentclass[11pt]{article}

\usepackage{hyperref}
\usepackage{multirow}
\usepackage{booktabs}
\usepackage{tabularx}
\usepackage{stfloats}

\usepackage[]{acl}

\usepackage{times}
\usepackage{latexsym}
\usepackage{booktabs}
\usepackage{xspace}
\usepackage{color,soul}
\usepackage{balance}
\usepackage{paralist}
\usepackage{verbatim}
\usepackage[normalem]{ulem}

\usepackage[T1]{fontenc}

\usepackage[utf8]{inputenc}
\usepackage{txfonts}
\usepackage{microtype}

\usepackage{subcaption}
\usepackage{graphicx}
\usepackage{makecell}

\definecolor{cbaseline1}{HTML}{EFC767}

\definecolor{cbaseline2}{HTML}{BD92BC}
\definecolor{cours}{HTML}{6EB4FD}
\definecolor{cgroundtruth}{HTML}{99C893}

\newcommand{\datasetname}{\textsc{FairytaleQA}\xspace}

\newcommand{\eg}{\emph{e.g.,}\xspace}%

\title{It is AI's Turn to Ask Humans a Question: \\ Question-Answer Pair Generation for Children's Story Books}

\author{Bingsheng Yao$^{\dagger}$  \\ Rensselaer Polytechnic Institute 
\And Dakuo Wang$^{\dagger}$  \\ IBM Research 
\And Tongshuang Wu \\ University of Washington  
\AND Zheng Zhang \\ University of Notre Dame 
\And Toby Jia-Jun Li \\ University of Notre Dame 
\AND Mo Yu$^{*}$ \\ WeChat AI, Tencent 
\And Ying Xu
\thanks{$^{\dagger}$ Equal contributions from the first authors: \texttt{yaob@rpi.edu, dakuo.wang@ibm.com; Work was done while Mo was at IBM Research.} $^*$ Corresponding authors.} \\ University of California Irvine}

\begin{document}
\maketitle
\begin{abstract}

Existing question answering (QA) techniques are created mainly to answer questions asked by humans. But in educational applications, teachers often need to decide what questions they should ask, in order to help students to improve their narrative understanding capabilities. We design an automated question-answer generation (QAG) system for this education scenario: given a story book at the kindergarten to eighth-grade level as input, our system can automatically generate QA pairs that are capable of testing a variety of dimensions of a student's comprehension skills.
Our proposed QAG model architecture is demonstrated using a new expert-annotated \datasetname dataset, which has 278 child-friendly storybooks with 10,580 QA pairs. Automatic and human evaluations show that our model outperforms state-of-the-art QAG baseline systems. On top of our QAG system, we also start to build an interactive story-telling application for the future real-world deployment in this educational scenario.

\end{abstract}

\section{Introduction}

%

\begin{table}[ht!]
\centering
\resizebox{.48\textwidth}{!}{%
\begin{tabular}{p{.48\textwidth}} 
\toprule
\textbf{\datasetname Dataset Source (Section) }                             \\ 
\small{Maie sighed. she knew well that her husband was right, but she could not give up the idea of a cow. the buttermilk no longer tasted as good as usual in the coffee;  }                                                              \\
\small{... ...}                                                             \\
\small{they were students, on a boating excursion, and wanted to get something to eat.'bring us a junket, good mother,' cried they to Maie.'ah! if only i had such a thing!' sighed Maie.}                                                \\ 

\midrule

\textcolor{cgroundtruth}{ \textbf{Ground-Truth}  }                                 \\ 
\begin{tabular}{@{\labelitemi\hspace{\dimexpr\labelsep+0.5\tabcolsep}}p{.45\textwidth}l@{}}\small{\textcolor{darkblue}{ \textbf{Q}}: What did the three young men ask for?~ ~ ~ ~ }\end{tabular}                                          \\
\begin{tabular}{@{\labelitemi\hspace{\dimexpr\labelsep+0.5\tabcolsep}}p{.43\textwidth}l@{}}\small{\textcolor{orange}{ \textbf{A}}: A junket. }\end{tabular} \\ 

\midrule

\textcolor{cbaseline1}{ \textbf{\texttt{2-Step Baseline}} }  \citep{shakeri2020end}         \\ 
\begin{tabular}{@{\labelitemi\hspace{\dimexpr\labelsep+0.5\tabcolsep}}p{.43\textwidth}l@{}}\small{\textcolor{darkblue}{ \textbf{Q}}: Why no more buttermilk for her husband to make?}\end{tabular}                                          \\
\begin{tabular}{@{\labelitemi\hspace{\dimexpr\labelsep+0.5\tabcolsep}}p{.43\textwidth}l@{}}\small{\textcolor{orange}{ \textbf{A}}: She could not give up the idea of a cow.}\end{tabular}                                                \\ 

\midrule

\textcolor{cbaseline2}{ \textbf{\texttt{PAQ} Baseline} } \citep{lewis2021paq}            \\ 
\begin{tabular}{@{\labelitemi\hspace{\dimexpr\labelsep+0.5\tabcolsep}}p{.43\textwidth}l@{}}\small{\textcolor{darkblue}{ \textbf{Q}}: What did maie think of when she thought of buttermilk?}\end{tabular}                                    \\
\begin{tabular}{@{\labelitemi\hspace{\dimexpr\labelsep+0.5\tabcolsep}}p{.43\textwidth}l@{}}\small{\textcolor{orange}{ \textbf{A}}: Sweet cream and fresh butter.
}\end{tabular}                                                              \\

\midrule

\textcolor{cours}{ \textbf{Our System} }                                     \\ 
\begin{tabular}{@{\labelitemi\hspace{\dimexpr\labelsep+0.5\tabcolsep}}p{.43\textwidth}l@{}}\small{\textcolor{darkblue}{ \textbf{Q}}: Why did the three young men want a junket?}\end{tabular}                                                \\
\begin{tabular}{@{\labelitemi\hspace{\dimexpr\labelsep+0.5\tabcolsep}}p{.43\textwidth}l@{}}\small{\textcolor{orange}{ \textbf{A}}: They wanted to get something to eat.
}\end{tabular}                                                              \\

\bottomrule
\end{tabular}
}
\caption { \small A sample of \datasetname story section as input and the QA pairs generated by human education experts, \texttt{2-step baseline} model,  \texttt{PAQ} baseline, and our QAG System. }
\vspace{-1em}
\label{tab:sample_generations}
\vspace{-10pt}
\end{table}



There has been substantial progress in the development of state-of-the-art (SOTA) question-answering (QA) models in the natural language processing community in recent years~\cite{xiong2019pretrained, karpukhin2020dense, cheng2020probabilistic, mou2021narrative}.
However, the opposite of QA tasks--question-answer generation (QAG) tasks that generate questions based on input text--is yet underexplored.
We argue, being able to ask a reasonable question is also an important indicator whether the reader comprehends the document, thus belongs to the reading comprehension(RC) task family.
QAG also contributes to important real-world applications, such as building automated systems to support teachers to efficiently construct assessment questions (and its correct answer) for the students at a scale~\cite{xu2021same,snyder2005assessment}.

Similar to training QA models, QAG model training requires high-quality and large-scale RC datasets (e.g., NarrativeQA~\cite{kovcisky2018narrativeqa}). 
However, many of the existing datasets are either collected via crowd-sourcing \citep{rajpurkar2016squad,kovcisky2018narrativeqa,reddy2019coqa}, or using automated retrievers \citep{nguyen2016ms,joshi2017triviaqa,dunn2017searchqa,kwiatkowski2019natural}, thus risking the quality and validity of labeled QA-pairs.
This risk becomes especially problematic when building applications in the education domain:
While existing QA models may perform well for the general domain, they fall short in understanding what are the most \textit{useful} QA pairs to generate for educational purposes.
Specifically, RC is a complex skill vital for children's achievement ~\cite{snyder2005assessment}, 
the datasets should contain questions that focus on a well-defined construct (e.g., narrative comprehension) and measure a full coverage of sub-skills within this construct (e.g., reasoning causal relationship and understanding emotion within narrative comprehension) using items of varying difficulty levels (e.g., inference making and information retrieval)~\cite{paris2003assessing}. 


In this work, we aim to develop a QAG system to generate high-quality QA-pairs, emulating how a teacher or parent would ask children when reading stories to them ~\cite{xu2021same}. 
Our system is built on a novel dataset that was recently released, \datasetname ~\cite{xu2022fairytaleqa}. This dataset focuses on narrative comprehension for elementary to middle school students and contains 10,580 QA-pairs from 278 narrative text passages of classic fairytales. As reported in ~\citet{xu2022fairytaleqa}, \datasetname was annotated by education experts and includes well-defined and validated narrative elements laid out in the education research \citep{paris2003assessing}, making it particularly appealing for RC research in the education domain. 


Our QAG system design consists of a three-step pipeline:
(1) to extract candidate answers from the given storybook passages through carefully designed heuristics based on a pedagogical framework; 
(2) to generate appropriate questions corresponding to each of the extracted answers using a state-of-the-art (SOTA) language model; and
(3) to rank top QA-pairs with a specific threshold for the maximum amount of QA-pairs for each section.

We compare our QAG system with two existing SOTA QAG systems: a \texttt{2-step baseline} system \cite{shakeri2020end} fine-tuned on \datasetname, and the other is an end-to-end generation system trained on a large-scale automatically generated RC dataset (\texttt{PAQ}) \cite{lewis2021paq}. We evaluate the generated QA-pairs in terms of similarity by Rouge-L precision score with different thresholds on candidate QA-pair amounts and semantic as well as syntactic correctness by human evaluation. We demonstrate that our QAG system performs better in both automated evaluation and human evaluation. Table~\ref{tab:sample_generations} is a sample of \datasetname story as input and the QA pairs generated by human education experts, \texttt{2-step baseline} model,  \texttt{PAQ} baseline, and our QAG System.

We conclude the paper by demoing an interactive story-telling application that built upon our QAG system to exemplify the applicability of our system in a real-world educational setting.

\section{Related Work}

\subsection{General QA Datasets}
There exists a large number of datasets available for narrative comprehension tasks. These datasets were built upon different knowledge resources and went through various QA-pair creating approaches. For instance, some focus on informational texts such as Wikipedia and website articles(\citet{rajpurkar2016squad}, \citet{nguyen2016ms}, \citet{dunn2017searchqa}, \citet{kwiatkowski2019natural}, \citet{reddy2019coqa}). Prevalent QA-pair generating approaches include crowd-sourcing \citep{rajpurkar2016squad,kovcisky2018narrativeqa,reddy2019coqa}, using automated QA-pair retriever \citep{nguyen2016ms,joshi2017triviaqa,dunn2017searchqa,kwiatkowski2019natural}, and etc. 
Datasets created by the approaches mentioned above are at risk of not consistently controlling the quality and validity of QA pairs due to the lack of well-defined annotation protocols specifically for the targeting audience and scenarios. Despite many of these datasets involving large-scale QA pairs, recent research \citep{kovcisky2018narrativeqa} found that the QA pairs in many RC datasets do not require models to understand the underlying narrative aspects. Instead, models that rely on shallow pattern matching or salience can already perform very well.  

NarrativeQA, for instance, \cite{kovcisky2018narrativeqa} is a large dataset with more than 46,000 human-generated QA-pairs based on abstractive summaries. Differing from most other RC datasets that can be answerable by shallow heuristics, the NarrativeQA dataset requires the readers to integrate information about events and relations expressed throughout the story content. Indeed, NarrativeQA includes a significant amount of questions that focus on narrative events and the relationship among events \citep{mou2021narrative}. 
One may expect that NarrativeQA could also be used for QAG tasks. In fact,
a couple of recent works use this dataset and train a network by combining a QG module and a QA module with a reinforcement learning approach\citep{tang2017question}. For example, \citet{wang2017joint} use the QA result to reward the QG module then jointly train the two sub-systems. In addition, \citet{nema2018towards} also explore better evaluation metrics for the QG system.
However, the NarrativeQA dataset is in a different domain than the educational context of our focus. Thus the domain adaptation difficulty is unknown.

\subsection{The \datasetname Dataset}
As previously mentioned, the general-purpose QA datasets (\eg SQuAD~\citep{rajpurkar2016squad}, MS MARCO~\citep{nguyen2016ms}) are unsuitable for children's education context, as they impose little structure on what comprehension skills are tested and heavily rely on crowd workers typically with limited education domain knowledge. \datasetname ~\cite{xu2022fairytaleqa} is a newly released RC dataset that precisely aims to solve those issues and complement the lack of a high-quality dataset resource for the education domain. This dataset contains over 10,000 high-quality QA-pairs from almost 300 children's storybooks, targeting students from kindergarten to eighth grade. 

As discussed in ~\citet{xu2022fairytaleqa}, \datasetname has two unique advantages that make it particularly useful for our project. First, the \datasetname was developed based on an evidence-based reading comprehension framework ~\cite{paris2003assessing}, which comprehensively focuses on seven narrative elements/relations contributing to reading comprehension: \texttt{character}, \texttt{setting}, \texttt{feeling}, \texttt{action}, \texttt{causal relationship}, \texttt{outcome resolution}, and \texttt{prediction} (Detailed definition and example of each aspect is described in Appendix~\ref{app:narrative_elements}). Second, the development of \datasetname followed a rigorous protocol and was fulfilled by trained annotators with educational research backgrounds. This process ensured that the annotation guideline was followed, the style of questions generated by coders was consistent, and the answers to the questions were factually correct. \datasetname was reported to have high validity and reliability through a validation study involving actual students ~\cite{xu2022fairytaleqa}.

\subsection{QAG Task}
A few years back, rule-based QAG systems \citep{heilman2009question,mostow2009generating,yao2010question,lindberg2013generating,labutov2015deep} were prevalent, but the generated QA suffered from the lack of variety. Neural-based models for question generation tasks~\cite{du2017learning, zhou2017neural, dong2019unified, scialom2019self,zhao2022storybookqag} have been an emerging research theme in recent years. 
But their focus are on the general domain QAG thus they only used the available general QA dataset for training, we have no idea how these models may perform in an education contxt.


In this paper, we use one recent work \citet{shakeri2020end} as our baseline. They proposed a two-step and two-pass QAG method that firstly generate questions (QG), then concatenates the questions to the passage and generates the answers in a second pass (QA). In addition, we include the recently-published Probably-Asked Questions (PAQ)~\cite{lewis2021paq} work as a second baseline. The PAQ system is an end-to-end QAG system trained on the PAQ dataset, a very large-scale QA dataset containing 65M automatically generated QA-pairs from Wikipedia. The primary issue with deep-learning-based models in the targeted children education application is that existing datasets and models do not consider the specific audience's language preference and the educational purposes~\cite{hill2015goldilocks,yao2012semantics}. 

Because both rule-based and neural-network-based approaches have their limitations inherently, in our work, we combine these two approaches to balance both the controllability of what types of QA pairs should be generated to better serve the educational purpose, and the diversity of the generated QA sequences.

\section{Pre-processing \datasetname Dataset}

\begin{table*}[ht]
\centering
\resizebox{.95\textwidth}{!}{%

\begin{tabular}{c||c|c|c|c||c|c|c|c||c|c|c|c} 
\toprule
\multirow{3}{*}{\textbf{\begin{tabular}[c]{@{}c@{}} \datasetname \\ Dataset \end{tabular}}}   &   \multicolumn{4}{c||}{\textbf{Train} }   &   \multicolumn{4}{c||}{\textbf{Validation} }   &   \multicolumn{4}{c}{\textbf{Test} }  \\
\cmidrule{2-13}
      &   \multicolumn{4}{c||}{232 Books with 8548 QA-pairs}   &   \multicolumn{4}{c||}{23 Books with 1025 QA-pairs}  
      &   \multicolumn{4}{c}{23 Books with 1007 QA-pairs}   \\ 
\cmidrule{2-13}       
      &   \textbf{Mean}   &    \textbf{S.D.}   &    \textbf{Min}   &    \textbf{Max}   &    \textbf{Mean}   &    \textbf{S.D.}   &    \textbf{Min}   &    \textbf{Max}   &    \textbf{Mean}   &    \textbf{S.D.}   &    \textbf{Min}   &    \textbf{Max}   \\

\midrule\midrule
\# section per story   &   14.4   &   8.8   &   2   &   60   &   16.5   &   10.0   &   4   &   43   &   15.8   &   10.8   &   2   &   55   \\ 
\# tokens per story   &   2160.9   &   1375.9   &   228   &   7577   &   2441.8   &   1696.9   &   425   &   5865   &   2313.4   &   1369.6   &   332   &   6330   \\ 
\# tokens per section   &   149.6   &   64.8   &   12   &   447   &   147.8   &   56.7   &   33   &   298   &   145.8   &   58.6   &   24   &   290   \\ 
\# questions per story   &   36.8   &   28.9   &   5   &   161   &   44.5   &   29.5   &   13   &   100   &   43.7   &   28.8   &   12   &   107   \\ 
\# questions per section   &   2.8   &   2.440   &   0   &   18   &   2.9   &   2.3   &   0   &   16   &      3.0   &   2.4   &   0   &   15   \\ 
\# tokens per question   &   10.2   &   3.2   &   3   &   27   &   10.9   &   3.2   &   4   &   24   &   10.5   &   3.1   &   3   &   25   \\ 
\# tokens per answer   &   7.1   &   6.0   &   1   &   69   &   7.7   &   6.3   &   1   &   70   &   6.8   &   5.2   &   1   &   44   \\
\bottomrule
\end{tabular}
}

\medskip
\vspace{-10pt}
\caption {\small Core statistics of the \datasetname dataset, which has 278 books and 10580 QA-pairs.}
\vspace{-1em}
\label{tab:stats_fairytale}
\end{table*}

\begin{figure*}[t]
    \centering
    \includegraphics[scale=0.56]{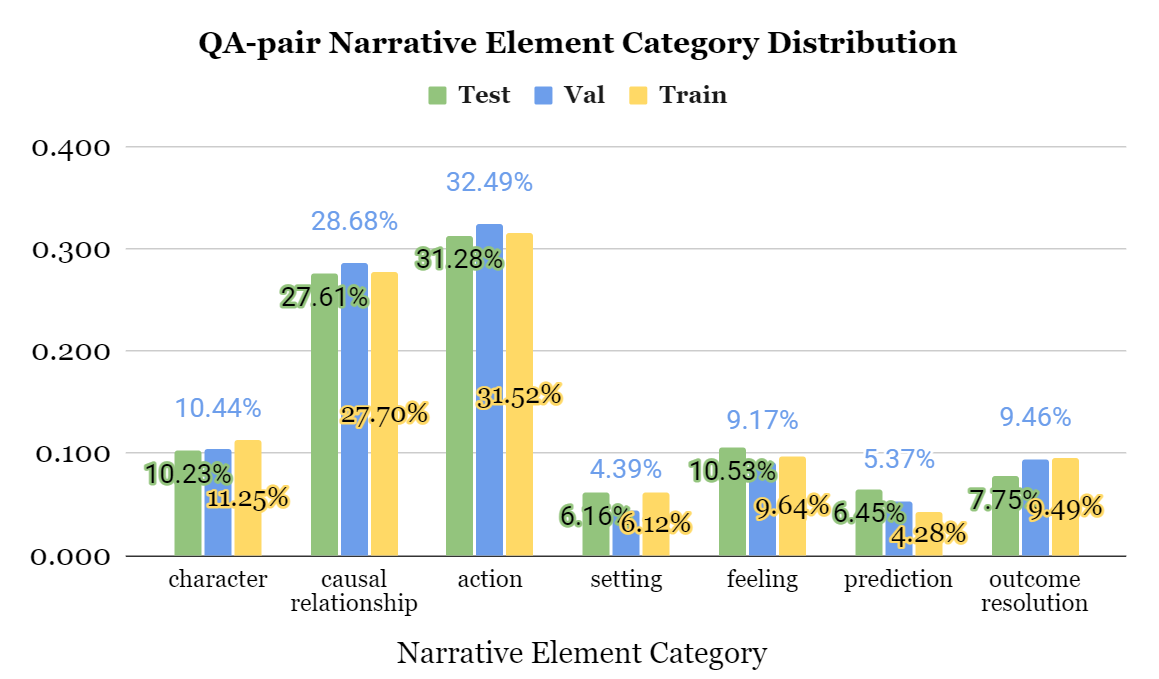}
    \caption{ \small Distribution of the QA-pairs belongs to each of the seven narrative element categories in the \datasetname dataset. }
    \vspace{-1em}
    \label{fig:distribution_fairytale}
\end{figure*}

The released \datasetname contained 10,580 QA-pairs from 278 books, and each question comes with a label indicating the narrative element(s)/relation(s) the question aims to assess. 

We split the dataset into train/validation/test splits with 232/23/23 books and 8,548/1,025/1,007 QA pairs. The split is random, but the statistical distributions in each split are consistent. Table~\ref{tab:stats_fairytale} shows core statistics of the \datasetname dataset in each split, and Figure~\ref{fig:distribution_fairytale} shows the distribution of seven types of annotations for the QA pairs across the three splits.



\section{Question Answer Generation System Architecture}

\begin{figure}[h]
    \centering
    \includegraphics[scale=0.5]{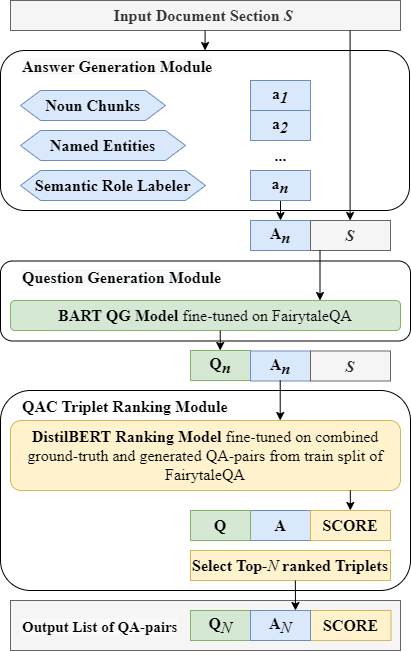}
    \caption{ \small QAG system design with three steps: rule-based answer extraction, NN-based question generation, and NN-based ranking. }
    \vspace{-1em}
    \label{fig:complete_pipeline}
\end{figure}



There are three sub-modules in our QA generation (QAG) pipeline: a heuristics-based answer generation module (AG), followed by a BART-based~\citep{lewis2019bart} question generation module (QG) module fine-tuned on \datasetname dataset, and a DistilBERT-based\citep{sanh2019distilbert} ranking module fine-tuned on \datasetname dataset to rank and select top $N$ QA-pairs for each input section. The complete QAG pipeline of our system is shown in Figure~\ref{fig:complete_pipeline}.


\subsection{Heuristics-based AG Module}
Based on our observation of the \datasetname dataset, educational domain experts seem to have uniform preferences over certain types of question and answer pairs (Figure~\ref{fig:distribution_fairytale}). 
This may be because these experts take the young children's learning objectives into consideration -- children's learning ability should be oriented toward specific types of answers to maximize their learning outcome. That is why educational experts rarely ask yes/no questions in developing or assessing children's reading comprehension.  
For automated QAG systems, we can design the system to mimic human behaviors either by defining heuristics rules for the answer extraction module, or leaving the filtering step to the end after the QA pairs are generated. However, the latter approach may have inherent risks that the training data could influence the types of answers generated. 

We decided to develop and apply the heuristic rules to the answer extraction module.
We observed that some narrative elements such as \texttt{characters}, \texttt{setting}, and \texttt{feelings} are mostly made up of name entities and noun chunks, for instance, the character name in a story, a particular place where the story takes place, or a specific emotional feeling. We then leverage the Spacy\footnote{\url{https://spacy.io/}} English model for Part-of-speech tagging on the input content to extract named entities and noun chunks as candidate answers to cover these three types of narrative elements.  

We further observed that the QA pairs created by education experts around the \texttt{action}, \texttt{causal relationship}, \texttt{prediction}, and \texttt{outcome resolution} categories are all related to a particular \textit{action event} in the story. Thus, the answers to these four types of questions are generally the description of the action event. We realize that Propbank's semantic roles labeler~\cite{palmer2005proposition} toolkit is constructive for extracting the action itself and the event description related to the action. 
We then leverage this toolkit to extract the trigger verb as well as other dependency nodes in the text content that can be put together as a combination of subject, verb, and object and use these as candidate answers for the latter four categories. 

Our answer extraction module can generate candidate answers that cover all 7 narrative elements with the carefully designed heuristics.

\begin{table}[t]
\centering
\resizebox{.48\textwidth}{!}{%
\small
\begin{tabular}{c||p{0.1\textwidth}||p{0.1\textwidth}}
\toprule
\multirow{2}{*}{\textbf{\begin{tabular}[l]{@{}c@{}} QG Models Comparison for \\  Our QAG System \end{tabular}}}  & \multicolumn{2}{c}{\textbf{Rouge-L} }\\
\cmidrule{2-3}
         &\hfil Validation &\hfil Test  \\
\midrule\midrule

\begin{tabular}[l]{@{}c@{}} BART fine-tuned on \\ NarrativeQA \end{tabular}   &\hfil   0.424   &\hfil   0.442   \\ 
\midrule

\begin{tabular}[l]{@{}c@{}} BART fine-tuned on \\ \datasetname \end{tabular}   &\hfil   \textbf{0.527}   &\hfil   \textbf{0.527}   \\ 
\midrule

\begin{tabular}[l]{@{}c@{}} BART fine-tuned on \\ NarrativeQA + \datasetname \end{tabular}   &\hfil   0.508   &\hfil   0.519   \\ 
\bottomrule
\end{tabular}
}

\caption{ \small Comparison on \datasetname dataset among QG models fine-tuned with different settings for the QG module of our QAG system. }
\vspace{-1em}

\label{tab:qg_performance}
\end{table}

\subsection{BART-based QG Module}

Following the answer extraction module that yields candidate answers, we design a QG module which takes a story passage and an answer as input, and generates the corresponding question as output. The QG task is basically a reversed QA task. Such a QG model could be either transfer-learned from another large QA dataset or fine-tuned on our \datasetname dataset. Mainstream QA datasets do cover various types of questions in order to comprehensively evaluate QA model's reading comprehension ability; for instance, NarrativeQA~\cite{kovcisky2018narrativeqa} is a large-scale QA corpus with questions that examine high-level abstractions to test the model's narrative understanding. 

We choose NarrativeQA dataset as an alternative option for fine-tuning our QG model because this dataset requires human annotators to provide a diverse set of questions about characters, events, etc., which is similar to the types of questions that education experts created for our \datasetname dataset. In addition, we leverage BART\cite{lewis2019bart} as the backbone model because of its superior performance on NarrativeQA according to the study in \cite{mou2021narrative}.

We perform a QG task comparison to examine the quality of questions generated for \datasetname dataset by one model fine-tuned on NarrativeQA, one on \datasetname, and the other on both the NarrativeQA and \datasetname. We fine-tune each model with different parameters and acquire the one with the best performance on the validation and test splits of \datasetname dataset. Results are shown in Table~\ref{tab:qg_performance}. We notice that the model fine-tuned on \datasetname alone outperforms the other methods. We attribute this to the domain and distribution differences between the two datasets.
That is why the model fine-tuned on both NarrativeQA and \datasetname may be polluted by the NarrativeQA training.
The best-performing model is selected for our QG module in the QAG pipeline.

\subsection{DistilBERT-based Ranking Module}

Our QAG system has generated hundreds of candidate QA-pairs through the first two modules. However, we do not know the quality of these generated QA-pairs by far, and it is unrealistic to send back all the candidate QA-pairs to users in a real-world scenario. Consequently, a ranking module is added to rank and select the top candidate QA-pairs, where the user is able to determine the upper limit of generated QA-pairs for each input text content. Here, the ranking task can be viewed as a classification task between the ground-truth QA-pairs created by education experts and the generated QA-pairs generated by our systems. 

We put together QA-pairs generated with the first two modules of our QAG system as well as ground-truth QA-pairs from the train/validation/test splits of \datasetname dataset, forming new splits for the ranking model, and fine-tuned on a pre-trained DistilBERT model. We test different input settings for the ranking module, including the concatenation of text content and answer only, as well as the concatenation of text content, question, and answer in various orders. Both input settings can achieve over $80\%$ accuracy on the test split, while the input setting of the concatenation of text content, question, and answer can achieve $F1=86.7\%$ with a leading more than $5\%$ over other settings. Thus, we acquire the best performing ranking model for the ranking module in our QAG system and allow users to determine the amount of top $N$ generated QA-pairs to be outputted.

\section{Evaluation}

We conduct both automated evaluation and human evaluation for the QAG task.  The input of the QAG task is a section of the story (may have multiple paragraphs), and the outputs are generated QA pairs. Unlike QA or QG tasks that each input corresponds to a single generated output no matter what model is used, the QAG task does not have a fixed number of QA-pairs to be generated for each section. Besides, various QAG systems will generate different amounts of QA-pairs for the same input content. Therefore, we carefully define an evaluation metric that is able to examine the quality of generated QA-pairs over a different amount of candidate QA-pairs. The comparison is on the validation and test splits of \datasetname.

\subsection{Automated Evaluation of QAG Task }

\subsubsection{Baseline QAG Systems}

We select a SOTA QAG system that uses a two-step generation approach ~\cite{shakeri2020end} as one baseline system (referred as \texttt{2-Step Baseline}). In the first step, it feeds a story content to a QG model to generate questions; then, it concatenates each question to the content passage and generates a corresponding answer through a QA model in the second pass. The quality of generated questions not only relies on the quality of the training data for the QG and QA models but also is not guaranteed to be semantically or syntactically correct because of the nature of neural-based models.

\begin{table}[t]
\centering
\resizebox{.48\textwidth}{!}{%
\small
\begin{tabular}{c||p{0.1\textwidth}||p{0.1\textwidth}}
\toprule
\multirow{2}{*}{\textbf{\begin{tabular}[l]{@{}c@{}} QA Models for \\  \texttt{2-Step Baseline} \end{tabular}}}  & \multicolumn{2}{c}{\textbf{Rouge-L} }\\
\cmidrule{2-3}
         &\hfil Validation &\hfil Test  \\
\midrule
\midrule
\begin{tabular}[l]{@{}c@{}} BART fine-tuned on \\ NarrativeQA \end{tabular}   &\hfil   0.475   &\hfil   0.492   \\ 
\midrule
\begin{tabular}[l]{@{}c@{}} BART fine-tuned on \\ \datasetname \end{tabular}   &\hfil   0.533   &\hfil   0.536   \\ 
\midrule
\begin{tabular}[l]{@{}c@{}} BART fine-tuned on \\ NarrativeQA + \datasetname \end{tabular}   &\hfil   \textbf{0.584}   &\hfil   \textbf{0.601}   \\ 
\bottomrule
\end{tabular}
}

\caption{ \small Comparison on \datasetname dataset among QA models fine-tuned with different settings for the \texttt{2-Step Baseline} system~\cite{shakeri2020end}. }
\vspace{-1em}

\label{tab:qa_performance}
\end{table}

We replicate this work by fine-tuning a QG model and a QA model on \datasetname dataset with the same procedures that help us select the best model for our QG module. We use pre-trained BART just like ours as the backbone model to ensure different model architectures do not influence the evaluation results. Unlike our QG module that takes both an answer and text content as the input, their QG model only takes the text content as input. Thus, we are not able to evaluate the QG model solely for this baseline. We replicate the fine-tuning parameters for our QG module to fine-tune the baseline QG model. For the selection of QA model used in the \texttt{2-Step Baseline}, similar to the QG experiments we present in Table~\ref{tab:qg_performance}, we fine-tune a pre-trained BART on each of the three settings: NarrativeQA only, \datasetname only, and both datasets. According to Table \ref{tab:qa_performance}, the model that fine-tuned on both NarrativeQA and \datasetname datasets performs much better than the other settings and outperforms the model that fine-tuned on \datasetname only by at least 6\%. We leverage the best performing QA model for the\texttt{ 2-Step Baseline} system.

In addition, we also include the recently published Probably-Asked Questions (\texttt{PAQ}) work as a second baseline system~\cite{lewis2021paq}. PAQ dataset is a semi-structured, very large scale Knowledge Base of 65M QA-pairs. \texttt{PAQ} system is an end-to-end QA-pair generation system that is made up of four modules: Passage Scoring, Answer Extraction, Question Generation, and Filtering Generated QA-pairs. The \texttt{PAQ} system is trained on the PAQ dataset. It is worth pointing out that during the end-to-end generation process, their filtering module requires loading the complete PAQ corpus into memory for passage retrieval, which leads us to an out-of-memory issue even with more than 50G RAM. \footnote{We do not use the filtering module for \texttt{PAQ} system because of unable to solve the memory issue with their provided code.} In comparison, our QAG system requires less than half of RAM in the fine-tuning process. In Table~\ref{tab:sample_generations}, we show a sample of \datasetname story section as input and the QA pairs generated by human education experts, \texttt{2-step baseline} model, \texttt{PAQ} baseline, and our QAG System. A few more examples are provided in Appendix~\ref{app:sample_generations}.

\subsubsection{Evaluation Metrics}

Since the goal of QAG is to generate QA-pairs that are most similar to the ground-truth QA-pairs given the same text content, we concatenate the question and answer to calculate the Rouge-L precision score for every single QA-pair evaluation. However, the amount of QA-pairs generated by various systems is different. It is unfair and inappropriate to directly compare all the generated QA-pairs from different systems. Moreover, we would like to see how QAG systems perform with different thresholds on candidate QA-pair amounts. In other words, we are looking at ranking metrics that given an upper bound $N$ as the maximum number of QA-pairs can be generated per section, how similar the generated QA-pairs are to the ground-truth QA-pairs. 

Generally, there are three different ranking metrics: Mean Reciprocal Rank (MRR), Mean Average Precision (MAP), and Normalized Discounted Cumulative Gain (NDCG). While MRR is only good to evaluate a single best item from the candidate list and NDCG requires complete rank ratings for each item, neither metric is appropriate in our case. As a result, We decide to use $MAP@N$, where $N \in [1, 3, 5, 10]$, as our evaluation metric for the QAG generation task. Furthermore, since the average amount of ground-truth answers are close to 3 per section in \datasetname dataset (Table~\ref{tab:stats_fairytale}), we expect the $MAP@3$ is the most similar to the actual use case, and we provide four $N$ to describe the comparison results and trends for QAG systems on the \datasetname. 

Here is the detailed evaluation process on $MAP@N$: for each ground-truth QA-pair, we find the highest Rouge-L precision score on the concatenation of generated question and answer, among top $N$ generated QA-pairs from the same story section. Then we average overall ground-truth QA-pairs to get the $MAP@N$ score. This evaluation metric evaluates the QAG system's performance on different candidate levels and is achievable even there is no ranking module in the system. For our QAG system, we just need to filter top $N$ QA-pairs from our ranking module; for the \texttt{2-Step Baseline} and the \texttt{PAQ} baseline system, we simply adjust a $topN$ parameter in the configuration.

\subsubsection{Evaluation Results}

Table \ref{tab:qag_performance} presents the evaluation results of our system and two SOTA baseline systems in terms of $MAP@N, N \in [1, 3, 5, 10]$. We observe our system outperforms both the \texttt{2-Step baseline} system and \texttt{PAQ} system in all settings with significantly better Rouge-L precision performance on both the validation and test splits of \datasetname dataset. According to the evaluation results, the \texttt{2-Step baseline} system suffers from the inherent lack of quality control of neural models over both generated answers and questions. We notice that the ranking module in our QAG system is an essential component of the system in locating the best candidate QA-pairs across different limits of candidate QA-pair amounts. The more candidate QA-pairs allowed to be selected for each section, the better our system performs compared to the other two baseline systems. Still, the Rouge-L score lacks the ability to evaluate the syntactic and semantic quality of generated QA-pairs. As a result, we further conduct a human evaluation to provide qualitative interpretations.

\begin{table*}[]
\centering
\resizebox{.95\textwidth}{!}{%

\small
\begin{tabular}{c||p{0.15\textwidth}||p{0.15\textwidth}||p{0.15\textwidth}||p{0.15\textwidth}}
\toprule
\multirow{3}{*}{\textbf{ QAG Systems }}  & \multicolumn{4}{c}{\textbf{MAP@N with Rouge-L Precision on Q+A for val/test splits} }\\
\cmidrule{2-5}
         &\hfil $N=10$ &\hfil $N=5$ &\hfil $N=3$ &\hfil $N=1$ \\
\midrule\midrule
Ours     &\hfil   \textbf{ 0.620 / 0.596 }   &\hfil   \textbf{ 0.543 / 0.523 }   &\hfil      \textbf{ 0.485 / 0.452 }   &\hfil   \textbf{ 0.340 / 0.310 }   \\
\midrule
 \texttt{2-Step} \texttt{Baseline}    &\hfil    0.443 / 0.422    &\hfil    0.370 / 0.353    &\hfil   0.322 / 0.305    &\hfil    0.225 / 0.216    \\
\midrule
\texttt{PAQ} \texttt{Baseline}     &\hfil    0.504 / 0.485    &\hfil    0.436 / 0.424    &\hfil    0.387 / 0.378   &\hfil    0.288 / 0.273    \\
\bottomrule
\end{tabular}

}
\caption{ \small Results of QAG task by our system and two baseline systems.  Left numbers are for validation split and right numbers are for test split.  }

\label{tab:qag_performance}
\end{table*}

\begin{table*}[ht]
\centering
\resizebox{.95\textwidth}{!}{%
\small
\begin{tabular}{c||p{0.1\textwidth}||p{0.1\textwidth}||p{0.1\textwidth}||p{0.1\textwidth}||p{0.1\textwidth}||p{0.1\textwidth}}
\toprule
  & \multicolumn{2}{c||}{\textbf{ Ours } }  &    \multicolumn{2}{c||}{\textbf{ \texttt{PAQ Baseline} } }    &   \multicolumn{2}{c}{\textbf{ \texttt{Groundtruth} } }    \\
\cmidrule{2-7}
   & \hfil   \textbf{M}   & \hfil   \textbf{SD}  & \hfil   \textbf{M}   & \hfil   \textbf{SD}   & \hfil   \textbf{M}   & \hfil   \textbf{SD}  \\

\midrule\midrule
\textbf{Readability** (1 to 5)}      &\hfil  4.71    &\hfil   0.70   &\hfil   4.08   &\hfil   1.13     &\hfil   4.95   &\hfil   0.28   \\
\midrule
\textbf{Question Relevancy* (1 to 5) }   &\hfil   4.39   &\hfil   1.15   &\hfil   4.18   &\hfil   1.22     &\hfil   4.92   &\hfil   0.33     \\
\midrule
Answer Relevancy (1 to 5)      &\hfil   3.99   &\hfil   1.51   &\hfil   3.90   &\hfil   1.62     &\hfil   4.83   &\hfil   0.57   \\
\bottomrule
\end{tabular}
}
\caption{ \small Human evaluation results.}
\label{table:human-eval}
\vspace{-1em}

\label{tab:human_performance}
\end{table*}





\subsection{Human Evaluation of QA Generation}
We recruited five human participants ($N=5$) to conduct a human evaluation to evaluate further our model generated QA quality against the ground-truth and the baseline (only against \texttt{PAQ} system as it outperforms the \texttt{2-Step Baseline}). 

In each trial, participants read a storybook section and multiple candidate QA pairs for the same section: three generated by the baseline \texttt{PAQ} system, three generated by our system (top-3), and the others were the ground-truth. 
Participants did not know which model each QA pair was from. The participant was asked to rate the QA pairs along three dimensions using a five-point Likert-scale. 

\begin{itemize}
\vspace{-5pt}
\item  \textit{Readability}: The generated QA pair is in readable English grammar and words.
\vspace{-10pt}
\item \textit{Question Relevancy}: The generated question is relevant to the storybook section.  
\vspace{-10pt}
\item \textit{Answer Relevancy}: The generated answer is relevant to the question.
\end{itemize}

We first randomly selected 7 books and further randomly selected 10 sections out of these 7 books (70 QA pairs).
Each participant was asked to rate these same 70 QA pairs to establish coding consistency.
The intercoder reliability score (Krippendoff's alpha~\cite{krippendorff2011computing}) among five participants along the four dimensions are between 0.73 and 0.79, which indicates an acceptable level of consistency.

Then, we randomly selected 10 books (5 from test and 5 from validation splits), and for each book, we randomly selected 4 sections. Each section, on average, has 9 QA-pairs (3 from each model). We assigned each section randomly to two coders. In sum, each coder coded 4 books (i.e. 16 sections and roughly 140 QA-pairs), and in total 722 QA-pairs were rated.

We conducted \textit{t-tests} to compare each model's performance. 
The result (Table \ref{table:human-eval}) shows that for the \textit{Readability} dimension, our model (avg=4.71, s.d.=0.70) performed significantly better than the \texttt{PAQ} model (avg=4.08, s.d.=1.13, $t(477)=7.33, p<.01$), but was not as good as the ground-truth (avg=4.95, s.d.=0.28, $t(479)=-4.85, p<.01$).

For the \textit{Question Relevancy} dimension, ground-truth also has the best rating (avg=4.92, s.d.=0.33), which was significantly better than the other two models. Our model (avg=4.39, s.d.=1.15) comes in second and outperforms baseline (avg=4.18, s.d.=1.22, $t(477)=1.98, p<.05$). The result suggests that questions generated by our model can generate more relevant to the story plot than those generated by the baseline model. 

For the \textit{Answer Relevancy} dimension, in which we consider how well the generated answer can answer the generated question, the ground-truth  (avg=4.83,s.d.=0.57) significant outperformed two models again. Our model (avg=3.99, s.d.=1.51) outperformed \texttt{PAQ} baseline model (avg=3.90, s.d.=1.62, $t(477)=0.58, p=.56$), but the result is not significant. 

All results show our model has above-average (>3) ratings, which suggests it reaches an acceptable user satisfaction along all three dimensions.

\begin{figure}[h!]
    \centering
    \includegraphics[width=0.6\columnwidth]{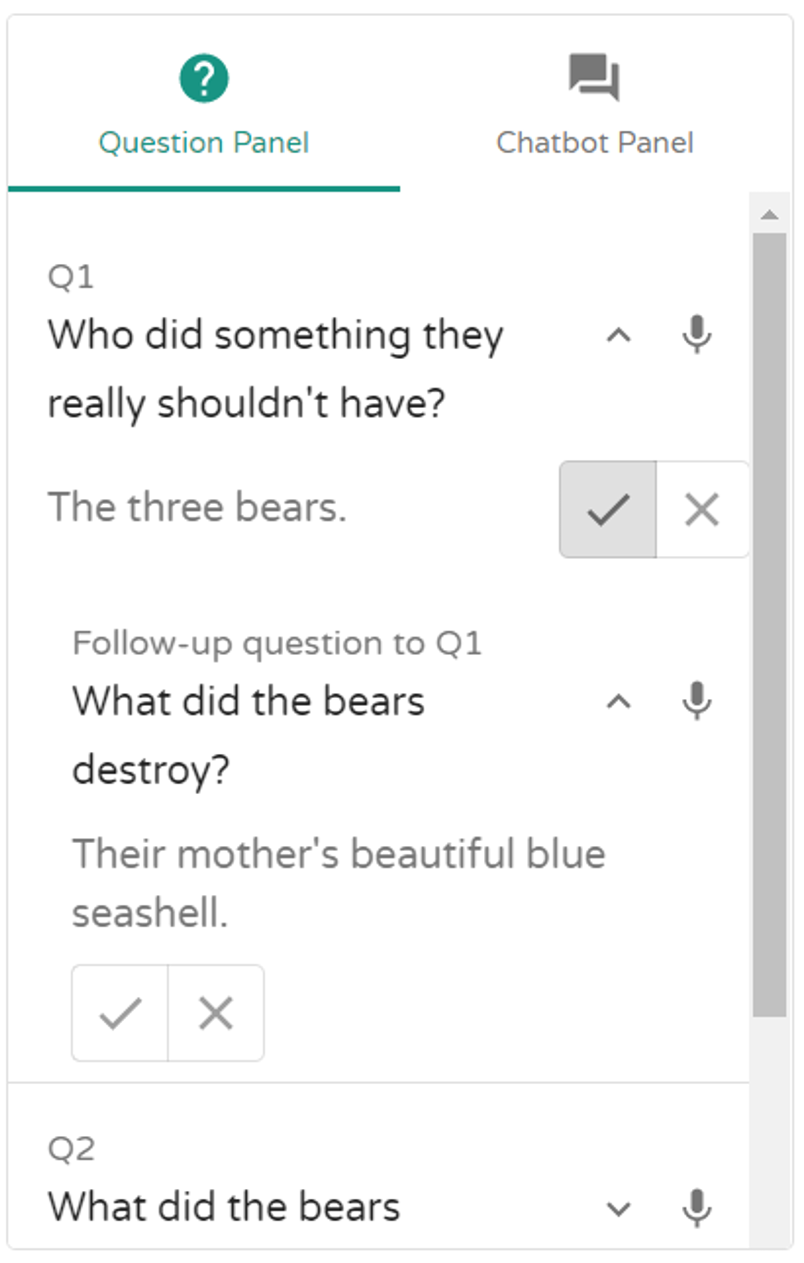}
    \caption{ \small The QA panel of our interactive storytelling application built upon our QAG system. The full user interface is shown in Appendix~\ref{app:interface}. }
    \vspace{-1em}
    \label{fig:demo_system_panel}
\vspace{-5pt}
\end{figure}

\subsection{Question Answer Generation in an Interactive Storytelling Application}

To exemplify the real-world application of our QAG system, we developed an interactive storytelling application built upon our QAG system. This system is designed to facilitate the language and cognition development of pre-school children via interactive QA activities during a storybook reading session. For example, as children move on to a new storybook page, the back-end QAG system will generate questions for the current section. Furthermore, to optimize child engagement in the QA session, the QAG system also generates follow-up questions for each answered question as shown in Figure~\ref{fig:demo_system_panel}. 
A conversational chatbot interacts with children, reads the story, facilitates questioning-and-answering via speech. The system can also keep track of child performance for the parents. 

A preliminary user study with 12 pairs of parents and children between the ages of 3-8 suggests that this application powered by our QAG system can successfully maintain engaging conversations with children about the story content. In addition, both parents and children found the system useful, enjoyable, and easy to use. Further evaluation and deployment details of this interactive storytelling system can be found in ~\cite{zhang_storybuddy:_2022}.

\section{Conclusion and Future Work}


In this work, we explore the question-answer pair generation task (QAG) in an education context for young children.  
Leveraging a newly-constructed expert-annotated QA dataset built upon child-oriented fairytale storybooks (\texttt{FairytaleQA}), we implemented a QA-pair generation pipeline which, as observed in human and automated evaluation, effectively supports our objective of automatically generating high-quality questions and answers at scale. 
To examine the model's applicability in the real world, we further built an interactive conversational storybook reading system that can surface the QAG results to children via speech-based interaction.

Our work lays a solid foundation for the promising future of using AI to automate educational question answering tasks. 
In the future, we plan to recruit educational experts to evaluate the educational efficacy of the QA-pairs as an additional evaluation dimension. 
Another future direction is to develop a context-aware multi-turn QAG system grounded by the story narratives (similar to \cite{li2021knowledge} ), where the generation of a new turn of QA is conditioned on previous generations as well as the book, so that it can enable new automated dialogue systems in the education setting. 



\section*{Acknowledgements}
This work was supported by the Rensselaer-IBM AI Research Collaboration (\url{http://airc.rpi.edu}), part of the IBM AI Horizons Network (\url{http://ibm.biz/AIHorizons}). 
The \datasetname dataset portion of the project is funded by Schmidt Futures.

\newpage

\bibliography{anthology,custom, anthology_1}
\bibliographystyle{acl_natbib}

\clearpage
\appendix

\section*{Appendix}

\begin{table*}[!htb]
\centering
\small
\begin{tabularx}{.95\textwidth}{p{0.1\textwidth}||X||X}
\toprule

\begin{tabular}[x]{@{}c@{}}\textbf{Narrative}\\\textbf{Element}\end{tabular} &  
\multicolumn{1}{c||}{\textbf{Definition} }  &
\multicolumn{1}{c}{\textbf{Example} }  \\

\midrule\midrule

Character   &   
Ask test takers to identify the character of the story or describe characteristics of characters  & 
\textcolor{red}{Q}: \textit{Who does Cassim marry after the death of their father?} \newline \textcolor{green}{A}: \textit{A wealthy woman}  \\

\midrule

Setting     &   
Ask about a place or time where/when story events take place and typically start with ``Where'' or ``When.''   &   
\textcolor{red}{Q}: \textit{Where did Lucdina and Jane Doll-cook buy their groceries?} \newline \textcolor{green}{A}: \textit{Ginger and Pickles}  \\
 
\midrule

Feeling     &   
Ask about the character's emotional status or reaction to certain events and are typically worded as ``How did/does/do …feel''  &   
\textcolor{red}{Q}: \textit{How did Ivan and Marie feel when Snowflake disappeared?} \newline \textcolor{green}{A}: \textit{sad}  \\

\midrule

Action      &   
Ask characters' behaviors or additional information about that behavior   &   
\textcolor{red}{Q}: \textit{What does Ali Baba do when his brother does not come back?} \newline \textcolor{green}{A}: \textit{goes to the cave to look for him}  \\

\midrule

Casual Relationship &   
Focus on two events that are causally related where the prior events have to causally lead to the latter event in the question. This type of question usually begins with ``Why'' or ``What made/makes.''      &   
\textcolor{red}{Q}: \textit{Why does Cassim forget the magic words to seal the cave?} \newline \textcolor{green}{A}: \textit{his greed and excitement over the treasures}  \\

\midrule

Outcome Resolution  &   
Ask for identifying outcome events that are causally led to by the prior event in the question. This type of question is usually worded as ``What happened/happens/has happened…after...''      &   
\textcolor{red}{Q}: \textit{What happened when January 1st came?} \newline \textcolor{green}{A}: \textit{There was still no money, and Pickles was unable to buy a dog license}  \\

\midrule

Prediction      &   
Ask for the unknown outcome of a focal event. This outcome is predictable based on the existing information in the text    &   
\textcolor{red}{Q}: \textit{What will happen to the Snow-man when the weather changes?} \newline \textcolor{green}{A}: \textit{thaw}  \\
 
\bottomrule
\end{tabularx}

\caption{ \small Definitions and examples for 7 narrative elements labeled in \datasetname Dataset }

\label{tab:narrative_elements}
\end{table*}

\begin{table*}[!ht]
\centering
\begin{tabularx}{.95\textwidth}{X||c|c||c|c||c|c}
\toprule
\multirow{2}{*}{\textbf{\begin{tabular}[c]{@{}c@{}} Category \end{tabular}}}   &   \multicolumn{2}{c||}{\textbf{Train} }   &   \multicolumn{2}{c||}{\textbf{Validation} }   &   \multicolumn{2}{c}{\textbf{Test} }  \\
\cmidrule{2-7}
      &   Count   &   Percentage   &   Count   &   Percentage   &   Count   &   Percentage   \\

\midrule\midrule
Character   &   962   &   0.112   &   107   &   0.104   &   103   &   0.102   \\ 
Causal Relationship   &   2368   &   0.277   &   294   &   0.286   &   278   &   0.276   \\ 
Action   &   2694   &   0.315   &   333   &   0.324   &   315   &   0.312   \\ 
Setting   &   523   &   0.061   &   45   &   0.043   &   62   &   0.061   \\ 
Feeling   &   824   &   0.096   &   94   &   0.091   &   106   &   0.105   \\ 
Prediction   &   366   &   0.0428   &   55   &   0.053   &   65   &   0.064   \\
Outcome Resolution   &   811   &   0.094   &   97   &   0.094   &   78   &   0.077   \\ 
\bottomrule

\end{tabularx}

\caption{ \small The number of QA-pairs belongs to each of the seven narrative element categories in the \datasetname dataset, inspired by ~\citep{paris2003assessing}.}
\vspace{-2em}
\label{tab:category_fairytale}
\end{table*}

\section{Definitions and examples for 7 narrative elements labeled in \datasetname Dataset}
\label{app:narrative_elements}

Table~\ref{tab:narrative_elements} shows detailed definition and example for each of the 7 narrative elements in \datasetname dataset.

\begin{table*}[!ht]
\centering
\resizebox{.95\textwidth}{!}{%

\begin{tabular}{p{.48\textwidth}} 
\toprule
\textbf{\datasetname Dataset Source (Section) }                             \\ 
\small{... ...}   \\
\small{ Then they passed through the dark cavern of horrors, when she'd have heard the most horrible yells, only that the fairy stopped her ears with wax. she saw frightful things, with blue vapours round them, and felt the sharp rocks and the slimy backs off rogs and snakes.when they got out of the cavern, they were at the mountain of glass; and then the fairy made her slippers so sticky with a tap of her rod that she followed the young corpse quite easily to the top. there was the deep sea a quarter of a mile under them, and so the corpse said to her,"go home to my mother, and tell her how far you came to do her bidding.farewell!" he sprung head-foremost down into the sea, and after him she plunged, without stopping a moment to think about it. }                \\
\small{... ...}   \\

\midrule

\textcolor{cgroundtruth}{ \textbf{Ground-Truth}  }                                 \\ 
\begin{tabular}{@{\labelitemi\hspace{\dimexpr\labelsep+0.5\tabcolsep}}p{.45\textwidth}l@{}}\small{\textcolor{darkblue}{ \textbf{Q}}: What did the fairy do to the youngest on the mountain of glass?~ ~ ~ ~ }\end{tabular}                                          \\
\begin{tabular}{@{\labelitemi\hspace{\dimexpr\labelsep+0.5\tabcolsep}}p{.43\textwidth}l@{}}\small{\textcolor{orange}{ \textbf{A}}: Made her slippers so sticky with a tap of her rod. }\end{tabular} \\ 

\midrule

\textcolor{cbaseline1}{ \textbf{\texttt{2-Step Baseline}} }  \citep{shakeri2020end}         \\ 
\begin{tabular}{@{\labelitemi\hspace{\dimexpr\labelsep+0.5\tabcolsep}}p{.43\textwidth}l@{}}\small{\textcolor{darkblue}{ \textbf{Q}}: What was the corpse doing?}\end{tabular}                                          \\
\begin{tabular}{@{\labelitemi\hspace{\dimexpr\labelsep+0.5\tabcolsep}}p{.43\textwidth}l@{}}\small{\textcolor{orange}{ \textbf{A}}: Go home to my mother, and tell her how far you came. }\end{tabular}     \\ 

\midrule

\textcolor{cbaseline2}{ \textbf{\texttt{PAQ} Baseline} } \citep{lewis2021paq}            \\ 
\begin{tabular}{@{\labelitemi\hspace{\dimexpr\labelsep+0.5\tabcolsep}}p{.43\textwidth}l@{}}\small{\textcolor{darkblue}{ \textbf{Q}}: What did the fairy stop her ears with? }\end{tabular}                                    \\
\begin{tabular}{@{\labelitemi\hspace{\dimexpr\labelsep+0.5\tabcolsep}}p{.43\textwidth}l@{}}\small{\textcolor{orange}{ \textbf{A}}: Wax. }\end{tabular}     \\

\midrule

\textcolor{cours}{ \textbf{Our System} }                                     \\ 
\begin{tabular}{@{\labelitemi\hspace{\dimexpr\labelsep+0.5\tabcolsep}}p{.43\textwidth}l@{}}\small{\textcolor{darkblue}{ \textbf{Q}}: What did the youngest princess see when she entered into the dark cavern of horrors? }\end{tabular}                                                \\
\begin{tabular}{@{\labelitemi\hspace{\dimexpr\labelsep+0.5\tabcolsep}}p{.43\textwidth}l@{}}\small{\textcolor{orange}{ \textbf{A}}: She saw frightful things , with blue vapours round them. }\end{tabular}        \\
\bottomrule
\end{tabular}

\begin{tabular}{p{.48\textwidth}} 
\toprule
\textbf{\datasetname Dataset Source (Section) }                             \\ 
\small{... ...} \\
\small{ Once upon a time there was a scholar, who wandered away from his home and went to emmet village. there stood a house which was said to be haunted. yet it was beautifully situated and surrounded by a lovely garden. so the scholar hired it. one evening he was sitting over his books, when several hundred knights suddenly came galloping into the room. they were quite tiny, and their horses were about the size of flies. they had hunting falcons and dogs about as large as gnats and fleas.they came to his bed in the corner of the room, and there they held a great hunt, with bows and arrows: one could see it all quite plainly.they caught a tremendous quantity of birds and game, and all this game was no larger than little grains of rice . }                \\
\small{... ...}   \\

\midrule

\textcolor{cgroundtruth}{ \textbf{Ground-Truth}  }                                 \\ 
\begin{tabular}{@{\labelitemi\hspace{\dimexpr\labelsep+0.5\tabcolsep}}p{.45\textwidth}l@{}}\small{\textcolor{darkblue}{ \textbf{Q}}: Who wandered away from his home and went to emmet village ?~ ~ ~ ~ }\end{tabular}                                          \\
\begin{tabular}{@{\labelitemi\hspace{\dimexpr\labelsep+0.5\tabcolsep}}p{.43\textwidth}l@{}}\small{\textcolor{orange}{ \textbf{A}}: A scholar. }\end{tabular} \\ 

\midrule

\textcolor{cbaseline1}{ \textbf{\texttt{2-Step Baseline}} }  \citep{shakeri2020end}         \\ 
\begin{tabular}{@{\labelitemi\hspace{\dimexpr\labelsep+0.5\tabcolsep}}p{.43\textwidth}l@{}}\small{\textcolor{darkblue}{ \textbf{Q}}: What happened one evening?}\end{tabular}          \\
\begin{tabular}{@{\labelitemi\hspace{\dimexpr\labelsep+0.5\tabcolsep}}p{.43\textwidth}l@{}}\small{\textcolor{orange}{ \textbf{A}}: Several hundred knights suddenly came galloping into the room . }\end{tabular}     \\ 

\midrule

\textcolor{cbaseline2}{ \textbf{\texttt{PAQ} Baseline} } \citep{lewis2021paq}            \\ 
\begin{tabular}{@{\labelitemi\hspace{\dimexpr\labelsep+0.5\tabcolsep}}p{.43\textwidth}l@{}}\small{\textcolor{darkblue}{ \textbf{Q}}: Where did the scholar go when he wandered away from home?}\end{tabular}   \\
\begin{tabular}{@{\labelitemi\hspace{\dimexpr\labelsep+0.5\tabcolsep}}p{.43\textwidth}l@{}}\small{\textcolor{orange}{ \textbf{A}}: Emmet village. }\end{tabular}     \\

\midrule

\textcolor{cours}{ \textbf{Our System} }                                     \\ 
\begin{tabular}{@{\labelitemi\hspace{\dimexpr\labelsep+0.5\tabcolsep}}p{.43\textwidth}l@{}}\small{\textcolor{darkblue}{ \textbf{Q}}: Who wandered away from his home and went to emmet village? }\end{tabular}       \\
\begin{tabular}{@{\labelitemi\hspace{\dimexpr\labelsep+0.5\tabcolsep}}p{.43\textwidth}l@{}}\small{\textcolor{orange}{ \textbf{A}}: A scholar. }\end{tabular}        \\
\bottomrule
\end{tabular}

}

\caption { \small Two more samples of \datasetname story as input and the QA pairs generated by human education experts, \texttt{2-step baseline} model,  \texttt{PAQ} baseline, and our QAG System. }
\vspace{-2em}
\label{tab:app_sample_generations}
\end{table*}

\section{Distribution of \datasetname annotations on 7 narrative elements}
\label{app:distribution}

Table~\ref{tab:category_fairytale} shows the distribution of QA-pair annotations on 7 essential narrative elements that are defined in \cite{paris2003assessing} of \datasetname dataset. The distribution of narrative elements is consistent across train/validation/test splits.

\section{QAG generation examples with 3 systems}
\label{app:sample_generations}
Table~\ref{tab:app_sample_generations} shows two more examples of \datasetname story section as input and the QA pairs generated by human education experts, \texttt{2-step baseline} model,  \texttt{PAQ} baseline, and our QAG System.

\section{User Interface of down-streaming application}
\label{app:interface}

\begin{figure*}[!htb]
    \centering
    \includegraphics[width=.95\textwidth]{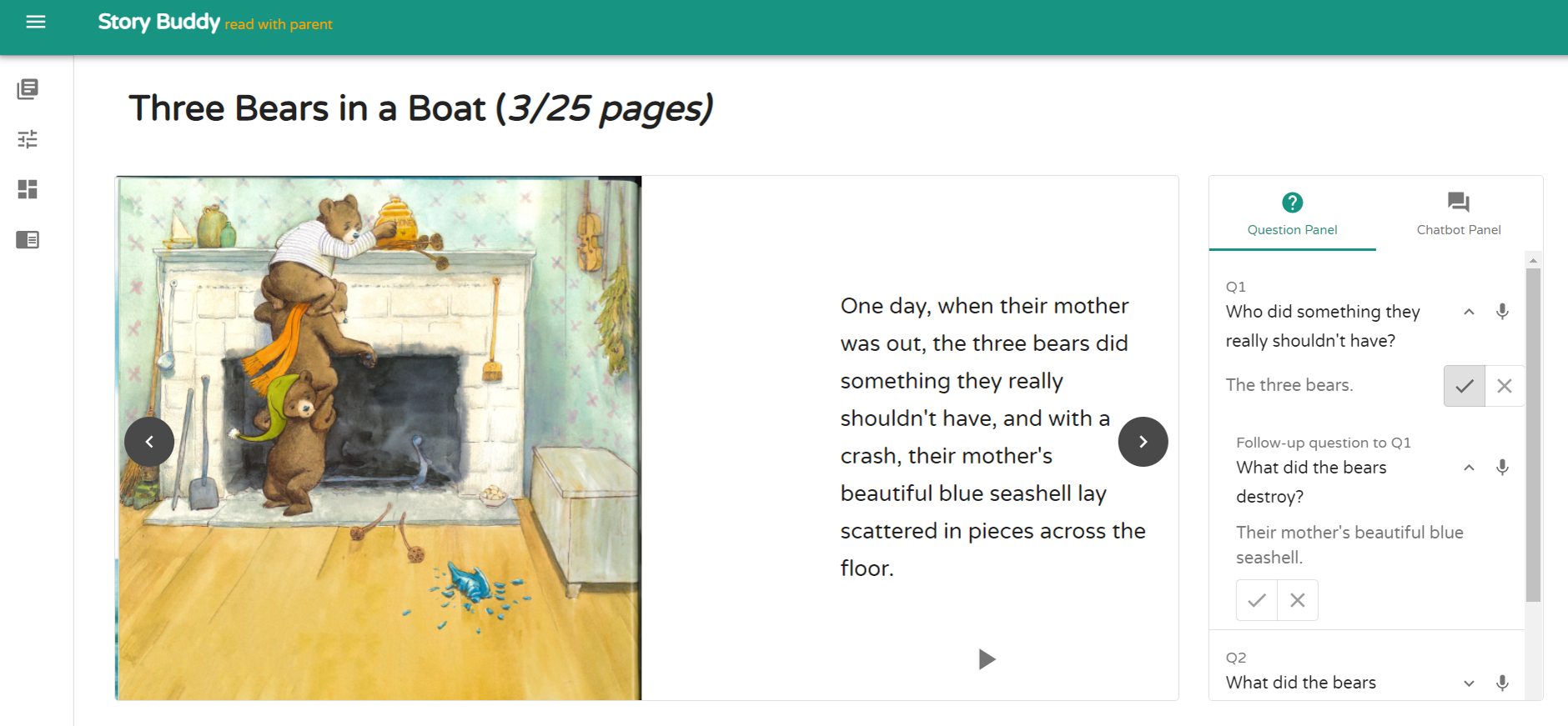}
    \caption{ \small The user interface of our down-streaming interactive storytelling system}
    \label{fig:demo_system}
    
\end{figure*}

Figure~\ref{fig:demo_system} is a screenshot of the interactive storytelling system interface \textsc{StoryBuddy}~\cite{zhang_storybuddy:_2022} for the down-streaming task of our QAG system in a real-world use scenario. Children can listen to the automatic story reading and try to answer the plot-relevant questions generated by the QAG system. They can answer the question via a microphone, and the system will judge the correctness of their answer. After answering a `parent' question, children can go further to answer a follow-up question or try out other `parent' questions. 

\section{Fine-tuning Parameters}
\label{appendix:parameters}
For fine-tuning the QA model for the \texttt{2-Step Baseline}, we select the best performing model with the following hyper-parameters: \textit{learning rate} = $5e^{-6}$; \textit{batch size} = $1$; \textit{epoch} = $1$. \\
For fine-tuning the QG model for our QAG system, we select the best performing model with the following hyper-parameters: \textit{learning rate} = $5e^{-6}$; \textit{batch size} = $1$; \textit{epoch} = $3$. 



\end{document}